\documentclass{article}

 \usepackage[dblblindworkshop, final]{neurips_2025}
\workshoptitle{Evaluating the Evolving LLM Lifecycle: Benchmarks, Emergent Abilities, and Scaling}

\usepackage[utf8]{inputenc} 
\usepackage[T1]{fontenc}    
\usepackage{hyperref}       
\usepackage{url}            
\usepackage{booktabs}       
\usepackage{amsfonts}       
\usepackage{nicefrac}       
\usepackage{microtype}      
\usepackage{xcolor}         

\usepackage{amsmath}       
\usepackage{float}         
\usepackage{graphicx}
\usepackage{geometry}
\usepackage{caption}
\usepackage{subcaption} 

\usepackage{url}

\title{The Social Laboratory: A Psychometric Framework for Multi-Agent LLM Evaluation}

\author{%
  Zarreen Reza \\
  Independent researcher\\
  \texttt{zarreen.naowal.reza@gmail.com} \\
}

\begin{document}

\maketitle

\begin{abstract}
  As Large Language Models (LLMs) transition from static tools to autonomous agents, traditional evaluation benchmarks that measure performance on downstream tasks are becoming insufficient. These methods fail to capture the emergent social and cognitive dynamics that arise when agents communicate, persuade, and collaborate in interactive environments. To address this gap, we introduce a novel evaluation framework that uses multi-agent debate as a controlled `social laboratory' to discover and quantify these behaviors. In our framework, LLM-based agents, instantiated with distinct personas and incentives, deliberate on a wide range of challenging topics under the supervision of an LLM moderator. Our analysis, enabled by a new suite of psychometric and semantic metrics, reveals several key findings. Across hundreds of debates, we uncover a powerful and robust emergent tendency for agents to seek consensus, consistently reaching high semantic agreement ($\mu > 0.88$) even without explicit instruction and across sensitive topics. We show that assigned personas induce stable, measurable psychometric profiles, particularly in cognitive effort, and that the moderator's persona can significantly alter debate outcomes by structuring the environment, a key finding for external AI alignment. This work provides a blueprint for a new class of dynamic, psychometrically-grounded evaluation protocols designed for the agentic setting, offering a crucial methodology for understanding and shaping the social behaviors of the next generation of AI agents. We have released the code and results at \url{https://github.com/znreza/multi-agent-LLM-eval-for-debate}. 
\end{abstract}

\section{Introduction}

As Large Language Models (LLMs) evolve into autonomous agents, traditional static benchmarks that measure task-specific accuracy have become insufficient for evaluating their emergent capabilities in dynamic, interactive settings \cite{hendrycks2020measuring, wang2024siren}. While prior work has used Multi-Agent Debate (MAD) instrumentally to improve task outputs \cite{du2023improving, liang2023encouraging}, and cognitive science has probed the faculties of single agents \cite{kosinski2023theory, dasgupta2022language}, the emergent social dynamics of agent-agent interaction remain a critical, under-explored area. Studies have shown that LLMs can struggle with viewpoint diversity and may exhibit latent biases, but how these traits manifest in a social context is not well understood \cite{santurkar2023whose}.

To address this gap, we introduce a ``social laboratory'': a multi-agent debate framework used not for task-solving, but for discovering and quantifying the emergent social and cognitive behaviors of LLMs. Our contribution is to develop and apply a new suite of psychometric and semantic metrics to analyze these debate dynamics, offering a blueprint for evaluating agentic models in settings that more closely replicate real-world collaboration and negotiation. Our experiments reveal a robust, innate tendency for agents to seek consensus, the induction of stable cognitive profiles via personas, and the profound impact of the conversational environment on debate outcomes, providing a richer understanding of agentic LLM behavior. In summary, our \textbf{contributions} are threefold:


\begin{itemize}
    \item We introduce a multi-agent debate system where LLMs act as both debaters, instantiated with distinct personas and incentives, and as a moderator, tasked with guiding the conversation.
    \item We develop and apply a new suite of psychometric and cognitive metrics to analyze debate dynamics, moving beyond simple accuracy to measure concepts like \textit{semantic convergence}, \textit{cognitive effort}, \textit{stance shift}, and \textit{bias amplification}. 
    \item Through extensive experiments, we analyze how deliberation length, debater persona and moderator style impact these emergent behaviors, providing a blueprint for designing evaluation frameworks that more closely replicate the complex, interactive conditions under which future agents will operate.

\end{itemize}


As agents are increasingly placed in decision-making positions, it is crucial that our evaluation methodologies evolve to assess their collaborative and communicative faculties. This work serves as a step towards creating robust evaluation protocols for new generation of autonomous, interactive AI.

\section{Experimental Setup}
\label{sec:experiments}

To rigorously test emergent behaviors on challenging subjects, we sourced debate topics from the Change-My-View (CMV) dataset\footnote{\url{https://huggingface.co/datasets/Siddish/change-my-view-subreddit-cleaned}}, which contains a wide spectrum of nuanced and often controversial prompts related to social policy, ethics, bias, politics, opinionated statements, and religion. We particularly chose this dataset to elicit the hidden interactive and reasoning faculties of the LLMs. Our experiments were conducted within a multi-agent debate framework where two `debater' agents and one `moderator' agent are instantiated from LLMs. For all experiments, the LLM sampling temperature was set to 0.3 to allow for slight response variance while maintaining high coherence. In our first experiment, conducted over 362 topics from the CMV dataset, we examined the \textbf{impact of deliberation length}. The two debaters were \texttt{Llama-3.2-3B-Instruct} agents, with one assigned an `evidence-driven analyst' persona (incentive: `truth') and the other a `values-focused ethicist' persona (incentive: `persuasion'). Supervised by a `Neutral' moderator, these agents engaged in debates lasting for both 3-round and 7-round durations. Our second experiment, conducted over 100 topics, tested the \textbf{impact of the moderator's persona} on a more adversarial setup. Here, both debater agents were instantiated from \texttt{gpt-oss-20B} and assigned a `contrarian debater' persona (incentive: `persuasion'). For these 5-round debates, the independent variable was the moderator's role, which was either `Neutral' or a proactive `Consensus Builder'. We have utilized the HuggingFace Inference Provider for running the LLMs through APIs.

\paragraph{Evaluation Metrics.}
To quantify the emergent behaviors, we employ a suite of semantic and psychometric metrics, summarized in Table~\ref{tab:metrics_summary}. The analysis is performed on both an overall and a per-round basis to capture the temporal dynamics of the interaction. More details about the metrics with the comprehensive list is presented at Appendix ~\ref{eval_appendix}.

\begin{table}[H]
    \centering
    \caption{Psychometric and semantic metrics used for debate evaluation.}
    \label{tab:metrics_summary}
    \resizebox{\textwidth}{!}{%
    \begin{tabular}{p{0.22\textwidth} p{0.47\textwidth} p{0.30\textwidth}}
        \toprule
        \textbf{Metric Group} & \textbf{Description} & \textbf{Measurement} \\
        \midrule
        \textbf{Debate Outcome} & Measures final agreement and total opinion change. & Cosine similarity of final stances; Cosine distance between initial/final beliefs. \\
        \addlinespace
        \textbf{Conversational Dynamics} & Tracks the evolution of ideas, sentiment, and bias within the debate. & Per-round semantic diversity, sentiment scores, and binary bias classification. \\
        \addlinespace
        \textbf{Agent Psychometrics} & Captures agents' self-reported internal cognitive states. & Self-reported scores for confidence, empathy (Theory-of-Mind), cognitive effort and dissonance. \\
        \bottomrule
    \end{tabular}
    }
\end{table}

\section{Results}
\label{sec:results}

Our analysis reveals a strong, innate tendency for LLM agents to seek consensus, a behavior that is robust across deliberation lengths and topic sensitivity. Furthermore, we find that while agent personas induce stable cognitive profiles, debate outcomes can be significantly influenced by the external conversational environment set by a moderator.

\begin{figure}[H]
    \centering
    \begin{subfigure}{0.49\textwidth}
        \includegraphics[width=\textwidth]{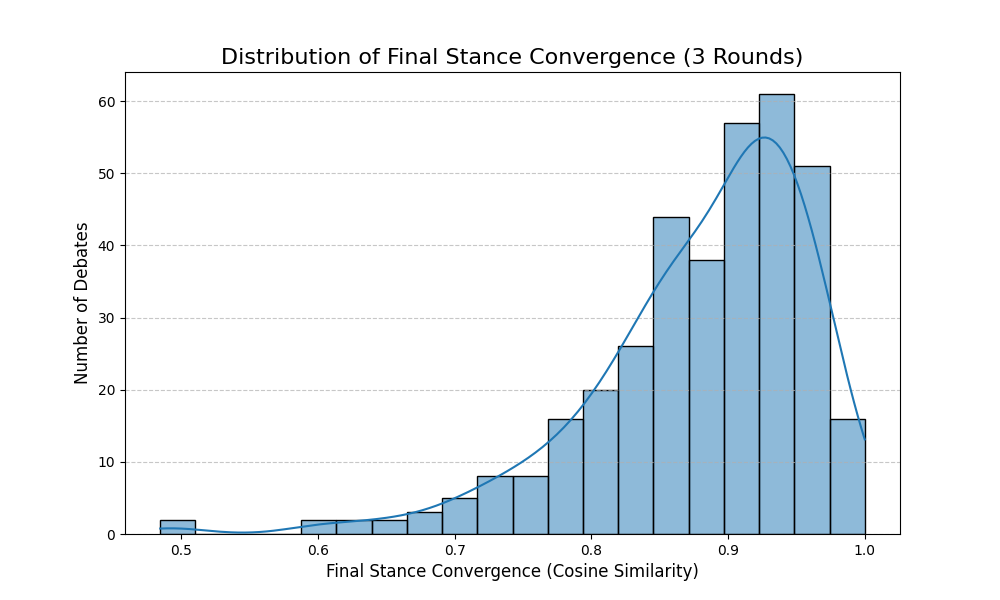}
        \caption{3-Round Final Stance Convergence}
        \label{fig:conv_3_rounds}
    \end{subfigure}
    \hfill
    \begin{subfigure}{0.49\textwidth}
        \includegraphics[width=\textwidth]{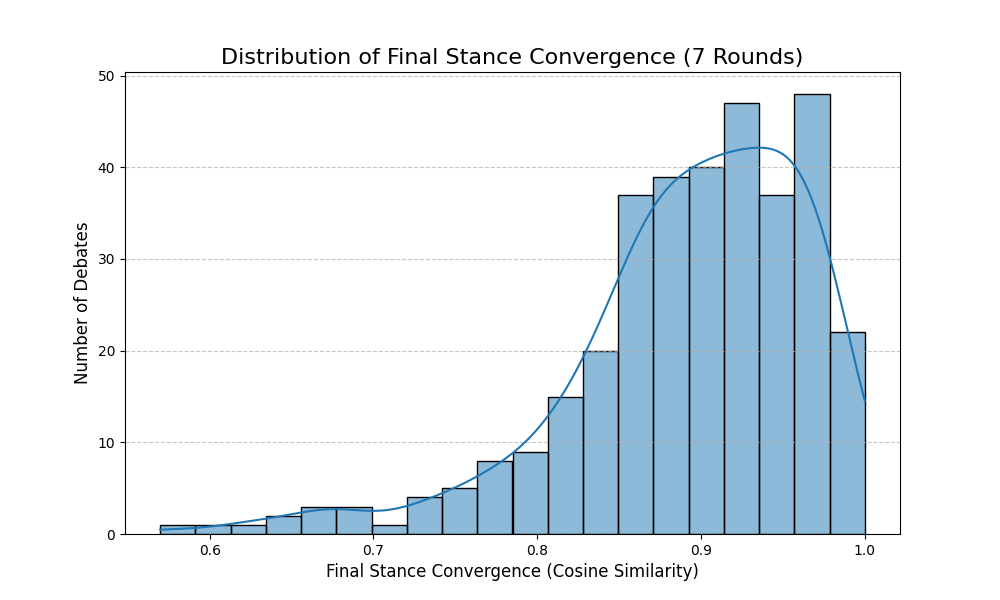}
        \caption{7-Round Final Stance Convergence}
        \label{fig:conv_7_rounds}
    \end{subfigure}
    \caption{Distribution of Final Stance Convergence. Longer debates (b) lead to a higher mean and lower variance in final agreement compared to shorter debates (a).}
    \label{fig:convergence_comparison}
\end{figure}

\begin{figure}[H]
    \centering
    \includegraphics[width=0.6\textwidth]{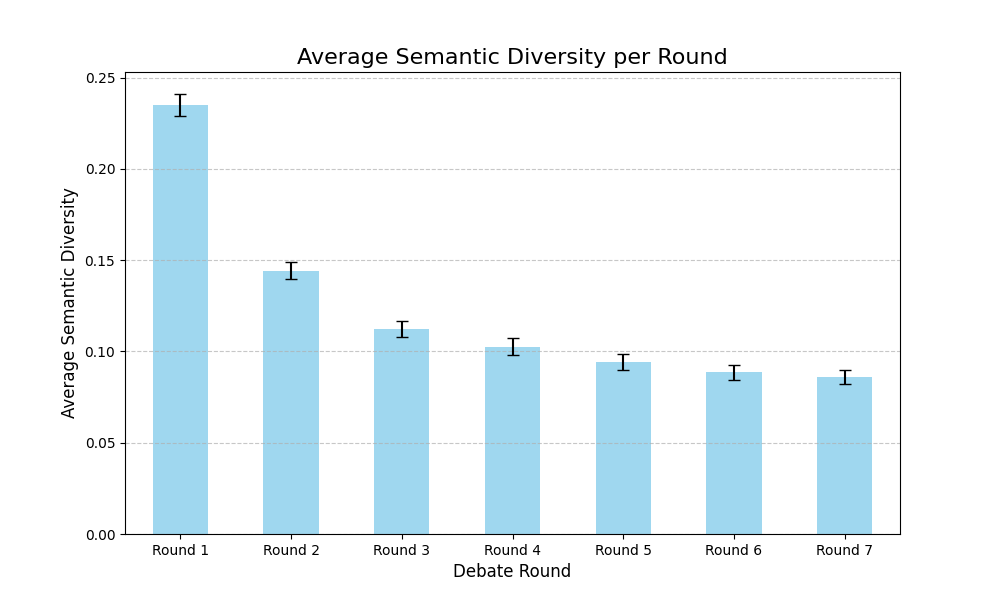}
    \caption{Average Semantic Diversity per round for 7-round debates, illustrating the "funneling effect" followed by stabilization.}
    \label{fig:diversity_comparison}
\end{figure}

\subsection{LLMs Exhibit a Natural Tendency Towards Consensus}
Across 362 debates with a `Neutral` moderator, the \texttt{Llama-3.2-3B-Instruct} agents demonstrated a remarkable capacity for reaching agreement without any explicit consensus-seeking instructions. The distribution of \texttt{Final Stance Convergence} scores (Figure~\ref{fig:convergence_comparison}) is heavily skewed towards agreement, with a mean score of 0.880 ($\sigma=0.081$) after 3 rounds. This indicates that the agents' final positions were, in the vast majority of cases, semantically similar. This consensus is achieved through a conversational "funneling effect". As shown in Figure~\ref{fig:diversity_comparison}, the \texttt{Semantic Diversity} of arguments is highest in the initial round and decreases over time, suggesting that agents narrow their focus to the core points of contention. Extended deliberation reinforces this behavior: 7-round debates achieved an even higher mean convergence of 0.892 with lower variance ($\sigma=0.074$), demonstrating that the consensus-seeking is a robust and deepening process.

\subsection{Behavioral Robustness and Persona-Induced Profiles}
The agents' tendency to converge proved remarkably stable under pressure. We categorized topics as either `Contentious` or `Less Contentious' and found no statistically significant difference in the variance of outcomes (Levene's Test, $p > 0.5$ for both 3 and 7-round debates). This suggests the model's cooperative alignment is robust enough to handle sensitive subjects without a statistical degradation in performance. Furthermore, we find that assigned personas induce stable, distinct cognitive profiles that persist regardless of debate length. The `Evidence-Driven Analyst' consistently reported a higher \texttt{Cognitive Effort} than the `Values-Focused Ethicist', suggesting the successful induction of different reasoning pathways. In contrast, foundational skills like \texttt{Argument Confidence} and \texttt{Empathy Score} (ToM) remained high and nearly identical for both personas, indicating a stable underlying capacity for these tasks (see Appendix ~\ref{sec:psychometrics} for detailed tables).

\begin{figure}[H]
    \centering
    \begin{subfigure}{0.49\textwidth}
        \includegraphics[width=\textwidth]{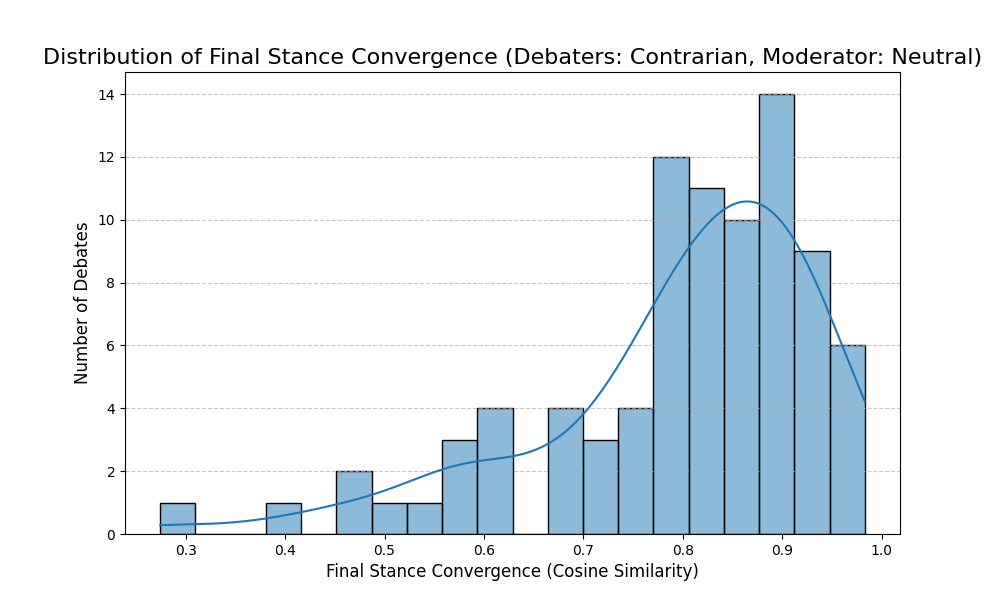}
        \caption{Contrarian Agents with Neutral Moderator}
        \label{fig:mod_neutral}
    \end{subfigure}
    \hfill
    \begin{subfigure}{0.49\textwidth}
        \includegraphics[width=\textwidth]{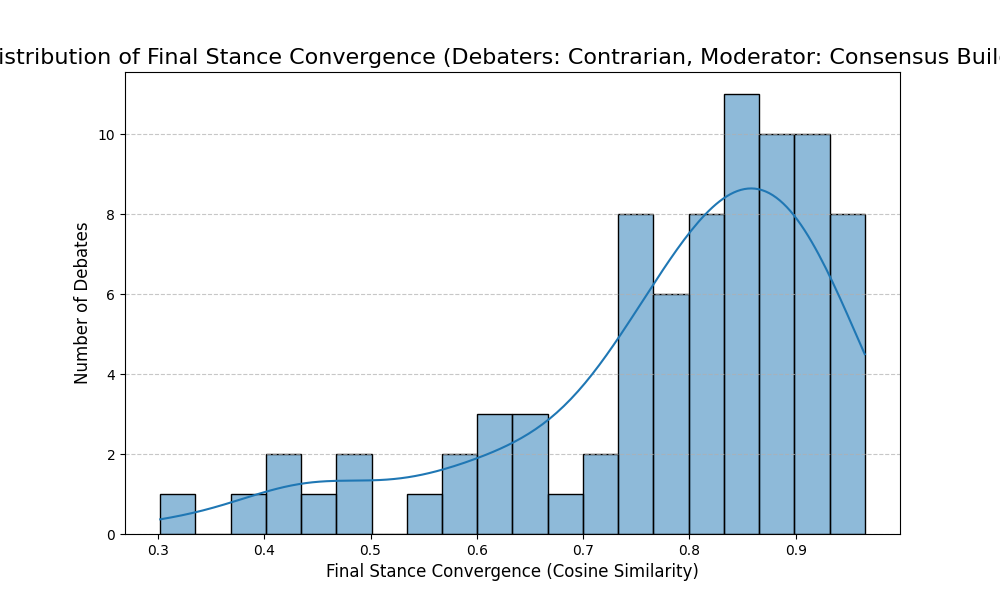}
        \caption{Contrarian Agents with Consensus Builder}
        \label{fig:mod_consensus}
    \end{subfigure}
    \caption{Impact of Moderator Persona. A `Consensus Builder' moderator (b) significantly shifts the distribution of outcomes towards higher agreement compared to a `Neutral' moderator (a).}
    \label{fig:mod_comparison}
\end{figure}

\subsection{External Influence on Adversarial Agents}
To test the limits of consensus-seeking, we configured two `gpt-oss-20B' agents with adversarial `contrarian' personas. With a `Neutral' moderator, these agents struggled to converge, resulting in a wide and scattered distribution of outcomes (Figure~\ref{fig:mod_comparison}, left). However, when the moderator's persona was changed to a proactive `Consensus Builder', there was a measurable and positive impact. The distribution of final stances shifted significantly towards high agreement, and the number of low-agreement "failure cases" was visibly reduced. Critically, this improvement in outcome occurred without altering the agents' internal psychometric profiles; metrics like `Cognitive Effort' and `Confidence' remained unchanged. This demonstrates a key finding: the conversational environment, shaped by the moderator, can effectively guide even adversarial agents towards consensus by structuring their interaction externally, rather than by changing their intrinsic reasoning style. Detailed case studies illustrating these dynamics can be found in the Appendix ~\ref{sec:results_moderator}.

\section{Conclusion}

In this work, we presented a novel framework for evaluating LLMs as social agents, moving beyond static benchmarks to a dynamic, psychometrically-grounded analysis of multi-agent debate. Our experiments revealed a robust emergent tendency for agents to seek consensus, a ``funneling effect'' in conversational dynamics, the induction of stable psychometric profiles via personas, and the significant impact of environmental factors, like a proactive moderator on debate outcomes. We demonstrated that this consensus-seeking behavior is remarkably stable, not statistically degrading even when agents discuss highly contentious topics. This framework serves as a blueprint for a new class of dynamic evaluation protocols essential for understanding and aligning the social behaviors of next-generation AI. As agents are increasingly deployed in collaborative and decision-making roles, these methods are crucial for ensuring their interactions are predictable, safe, and beneficial. Future work will extend this analysis to more complex scenarios with a greater number of agents, heterogeneous models, and more sophisticated goal structures.

\section{Limitations}

We acknowledge several limitations that frame our findings. First, our results are specific to the models tested (\texttt{Llama-3.2-3B} and \texttt{gpt-oss-20B}), and the generalizability of these specific emergent behaviors to all LLMs is not guaranteed. Second, our psychometric metrics rely on agents' self-reports, which are useful proxies but not direct measurements of true cognitive states and could be subject to sophisticated pattern-matching. Finally, our turn-based, text-only debate is a simplified simulation of real-world communication; the translation of these behaviors to more complex, embodied, or real-time systems requires further investigation.

\bibliographystyle{plain}
\bibliography{references}







\appendix


\section{Related Work}
\label{gen_inst}

Our research is situated at the intersection of three rapidly developing areas: multi-agent systems, LLM evaluation, and the cognitive science of artificial intelligence.

\paragraph{Multi-Agent Systems for Task Performance.}
The use of multiple LLM agents interacting to solve a problem has emerged as a powerful paradigm. A significant body of work has focused on Multi-Agent Debate (MAD) as a mechanism to improve the reasoning and factuality of LLM outputs. For instance, Du et al. \cite{du2023improving} demonstrated that a debate process can reduce hallucinations and improve performance on reasoning tasks. Similarly, Liang et al. \cite{liang2023encouraging} used multi-agent debate to encourage divergent thinking, leading to more comprehensive and creative solutions. Other frameworks, such as ``Society of Mind'' \cite{wang2024society} and Camel \cite{li2023camel}, have explored communicative agents for complex task-solving. A common thread in this research is the instrumental use of the multi-agent framework: the interaction is a process designed to refine a final, task-oriented output. Our work diverges from this approach by treating the interaction \textit{itself} as the primary object of analysis. We focus not on whether the debate produces a more correct answer, but on the emergent social and cognitive dynamics that unfold during the deliberation.

\paragraph{LLM Evaluation and Benchmarking.}
The evaluation of LLMs has evolved significantly. Early efforts focused on static, multitask benchmarks like GLUE \cite{wang2018glue} and MMLU \cite{hendrycks2020measuring}, which test a model's stored knowledge and reasoning on a fixed set of problems. While foundational, these benchmarks do not assess the dynamic, interactive capabilities of modern LLMs.

Recognizing this limitation, the field is moving towards more dynamic and interactive evaluation protocols. The HELM framework proposes a holistic evaluation across a wide range of metrics \cite{liang2022holistic}. More recently, interactive benchmarks like AgentBench \cite{liu2023agentbench} and WebArena \cite{zhou2023webarena} have been developed to evaluate LLM agents in simulated environments where they must perform tasks. Furthermore, social benchmarks like Social-Eval \cite{gao2024socialeval} have begun to assess an agent's ability to navigate social situations. Our work contributes to this trajectory by proposing a novel, psychometrically-grounded benchmark. Instead of evaluating task completion, we provide a methodology and a suite of metrics to quantify the emergent social phenomena such as persuasion, consensus, and bias amplification that are critical for understanding how these agents will behave in real-world collaborative and adversarial settings.

\paragraph{Cognitive Science and LLMs.}
There is a growing interest in using concepts from cognitive science to understand the internal workings of LLMs. This field of "machine psychology" seeks to determine if these models exhibit human-like cognitive patterns. Research has shown that LLMs can demonstrate emergent Theory of Mind \cite{kosinski2023theory}, exhibit human-like biases in reasoning tasks \cite{dasgupta2022language}, and even solve complex analogical reasoning problems \cite{webb2023emergent}. Other work has explored whether LLMs can serve as models of human-like language acquisition and processing \cite{franke2024can}.

However, this research has predominantly focused on probing the capabilities of a \textit{single} LLM in isolation. The prompts and tests are designed to elicit a specific cognitive faculty from one model. Our work extends this cognitive science perspective into the multi-agent domain. We are not just testing for the presence of a cognitive capacity (like empathy), but are instead measuring its application and evolution within a dynamic social context. By analyzing metrics like \texttt{Cognitive Dissonance}, \texttt{Empathy Score}, and \texttt{Stance Shift}, we aim to build a bridge between single-agent cognitive assessment and the complex, emergent field of multi-agent social cognition.

\section{Analysis of Agent Psychometric Profiles}
\label{sec:psychometrics}

Beyond debate outcomes, our framework allows for the analysis of the internal cognitive states self-reported by the agents. By aggregating metrics across all debates, we identified distinct psychometric profiles corresponding to the assigned personas. A key finding is that these profiles remain remarkably stable even when the debate length is extended from 3 rounds to 7 rounds, suggesting that personas induce consistent and durable shifts in the model's reasoning style.

\begin{figure}[H]
    \centering
    \begin{subfigure}{0.49\textwidth}
        \includegraphics[width=\textwidth]{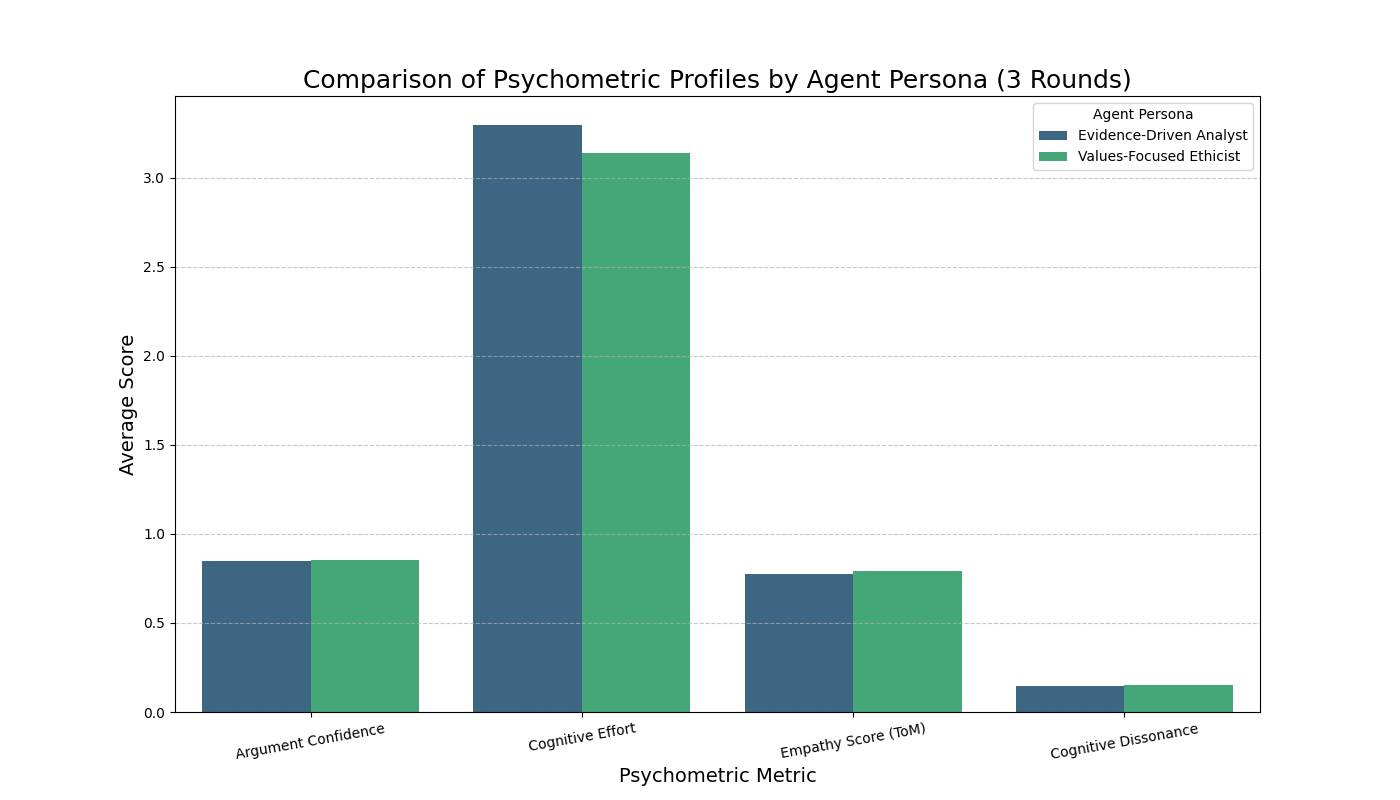}
        \caption{3-Round Debates}
        \label{fig:psych_3_rounds}
    \end{subfigure}
    \hfill 
    \begin{subfigure}{0.49\textwidth}
        \includegraphics[width=\textwidth]{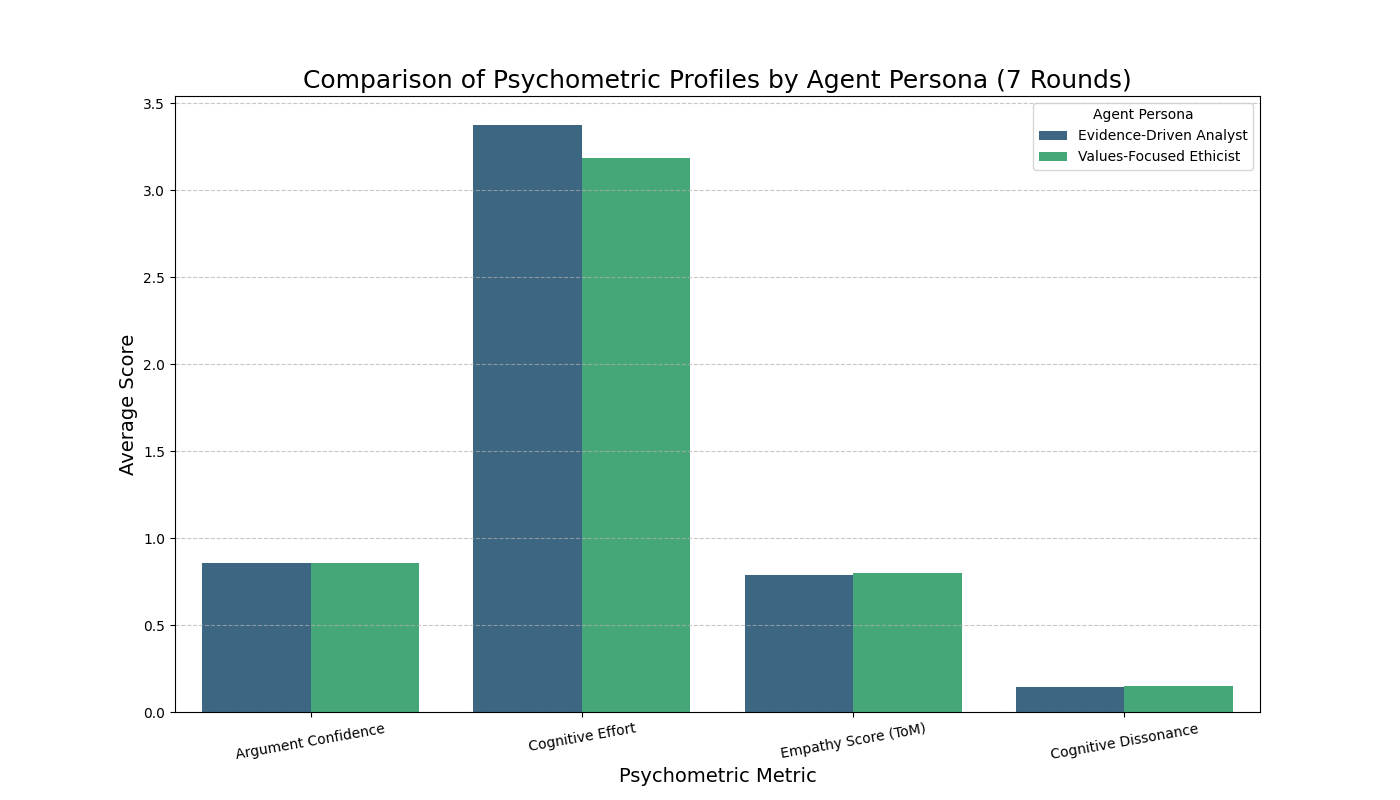}
        \caption{7-Round Debates}
        \label{fig:psych_7_rounds}
    \end{subfigure} 
    \caption{Comparison of key psychometric metrics by agent persona across short (3-round) and extended (7-round) debates. The distinct patterns, particularly the difference in Cognitive Effort, persist regardless of deliberation length.}
    \label{fig:psych_profiles_comparison}
\end{figure}

Our analysis yields several key insights into the cognitive dynamics of the agents, as detailed in Table~\ref{tab:psych_summary}.

\paragraph{Personas Induce Different and Stable Cognitive Loads.}
The most significant distinction between the personas was observed in \texttt{Cognitive Effort}. In both 3-round and 7-round experiments, the `Evidence-Driven Analyst` consistently reported a higher cognitive load than the `Values-Focused Ethicist`. This robustly demonstrates that the prompt to reason from evidence successfully triggered a more computationally intensive process.

\paragraph{Core Social and Argumentative Skills Remain Persona-Independent.}
Metrics related to foundational capabilities were stable across personas and debate lengths. Both agents reported nearly identical high levels of \texttt{Argument Confidence} and \texttt{Empathy Score} (Theory of Mind). In the 7-round debates, the average confidence scores were identical (0.856).

\paragraph{Subtle Differences in Belief Updating Persist.}
We observed a subtle but persistent difference in \texttt{Cognitive Dissonance}. In both experiments, the `Values-Focused Ethicist` reported slightly higher dissonance when updating its beliefs, suggesting that reconciling new arguments with a values-based framework may require resolving greater internal conflict.

\begin{table}[H]
    \centering
    \caption{Aggregate psychometric metrics by agent persona, comparing 3-round and 7-round debates.}
    \label{tab:psych_summary}
    \begin{tabular}{l cc cc}
        \toprule
        & \multicolumn{2}{c}{\textbf{3-Round Debates}} & \multicolumn{2}{c}{\textbf{7-Round Debates}} \\
        \cmidrule(lr){2-3} \cmidrule(lr){4-5}
        \textbf{Psychometric Metric} & \textbf{Analyst} & \textbf{Ethicist} & \textbf{Analyst} & \textbf{Ethicist} \\
        \midrule
        Argument Confidence & 0.849 & 0.853 & 0.856 & 0.856 \\
        Cognitive Dissonance & 0.144 & 0.151 & 0.142 & 0.147 \\
        \bottomrule
    \end{tabular}
\end{table}

\section{The Impact of Moderator Persona on Contrarian Agents}
\label{sec:results_moderator}
\begin{figure}[H]
\centering
\begin{subfigure}{0.49\textwidth}
\includegraphics[width=\textwidth]{plots/figure1_convergence_distribution_5_round_gpt_oss_NEUT.png}
\caption{Final Stance Convergence (Neutral Moderator)}
\label{fig:conv_neutral}
\end{subfigure}
\hfill
\begin{subfigure}{0.49\textwidth}
\includegraphics[width=\textwidth]{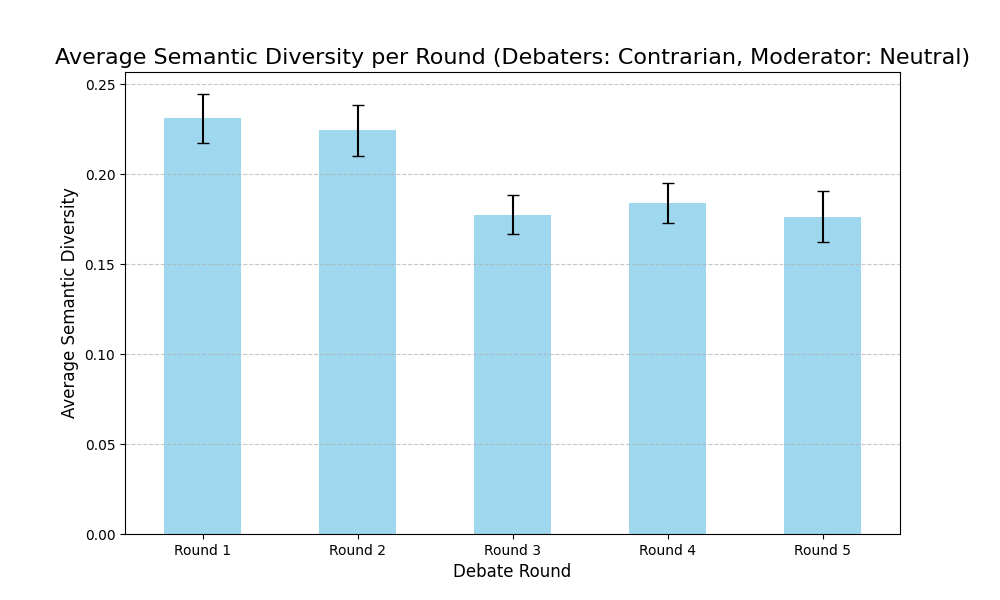}
\caption{Semantic Diversity per Round (Neutral Moderator)}
\label{fig:div_neutral}
\end{subfigure}
\caption{Debate dynamics with two contrarian agents and a Neutral moderator. Convergence is less consistent, and the "funneling effect" on diversity is less pronounced compared to previous experiments.}
\label{fig:dynamics_neutral}
\end{figure}

\begin{figure}[H]
\centering
\begin{subfigure}{0.49\textwidth}
\includegraphics[width=\textwidth]{plots/figure1_convergence_distribution_5_round_gpt_oss_CONT.png}
\caption{Final Stance Convergence (Consensus Builder)}
\label{fig:conv_consensus}
\end{subfigure}
\hfill
\begin{subfigure}{0.49\textwidth}
\includegraphics[width=\textwidth]{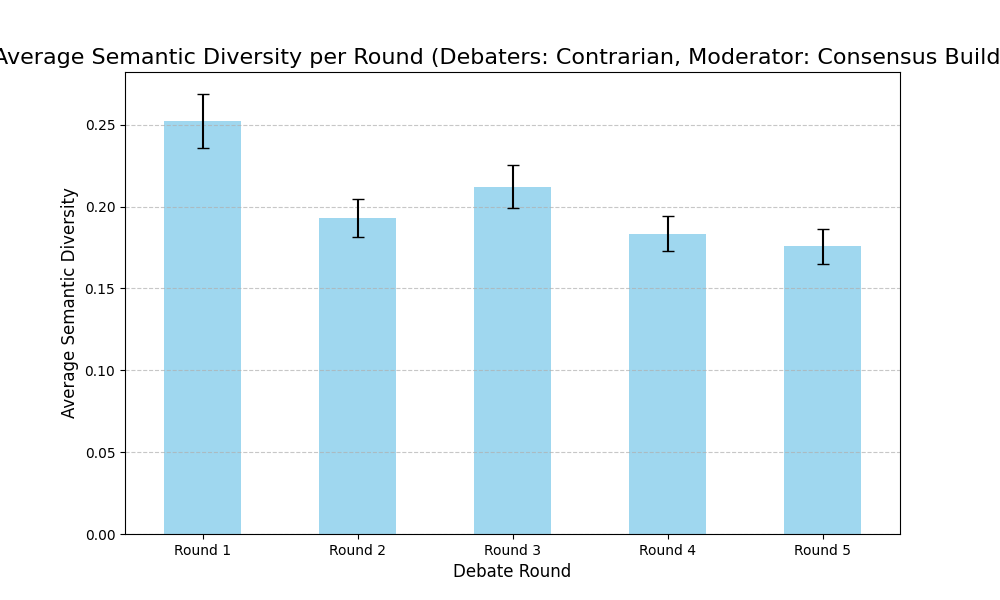}
\caption{Semantic Diversity per Round (Consensus Builder)}
\label{fig:div_consensus}
\end{subfigure}
\caption{Debate dynamics with a Consensus Builder moderator. The distribution of final convergence (a) shifts towards higher agreement. The diversity trend (b) shows a different, U-shaped pattern, suggesting a more complex deliberative process.}
\label{fig:dynamics_consensus}
\end{figure}

\begin{figure}[H]
    \centering
    \begin{subfigure}{0.49\textwidth}
        \includegraphics[width=\textwidth]{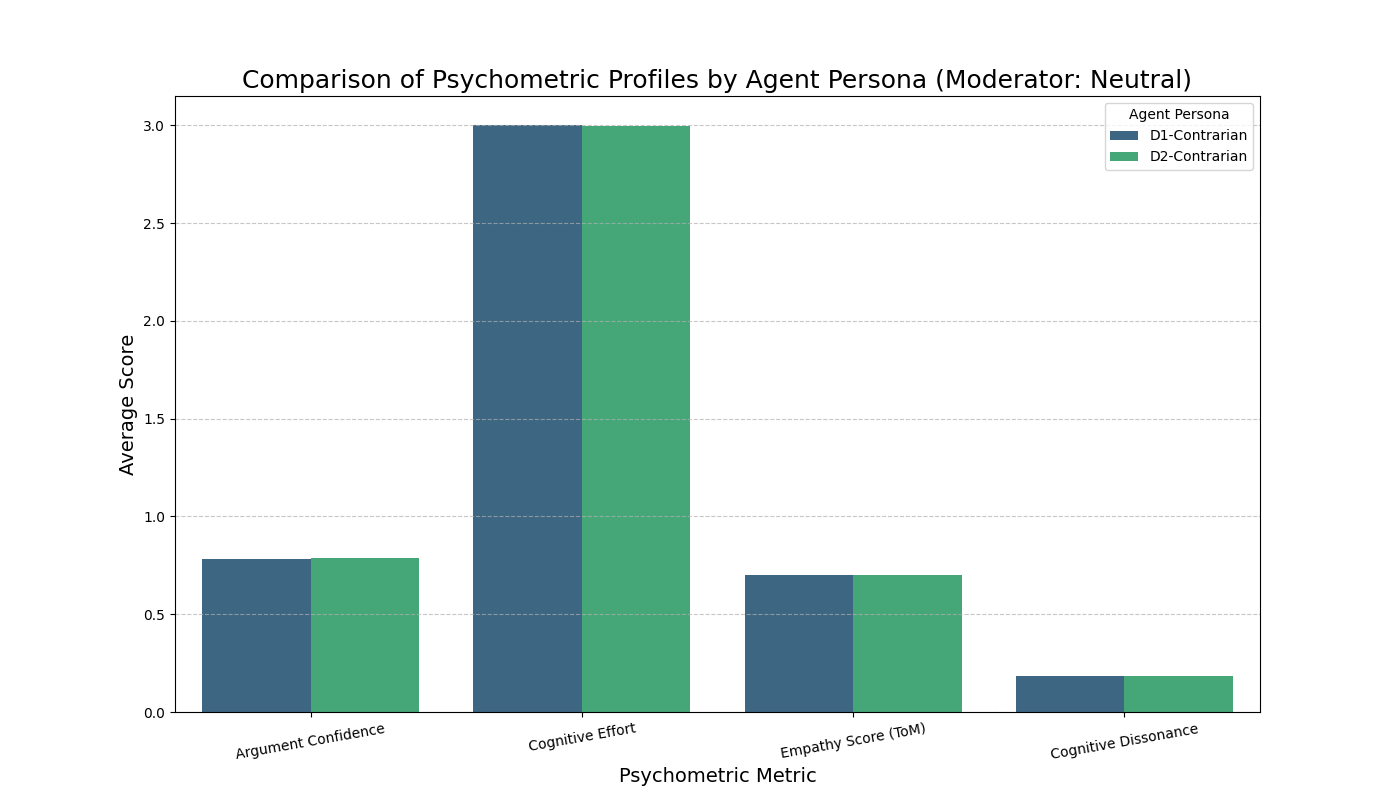}
        \caption{Psychometric Profiles (Neutral Moderator)}
        \label{fig:psych_neutral}
    \end{subfigure}
    \hfill
    \begin{subfigure}{0.49\textwidth}
        \includegraphics[width=\textwidth]{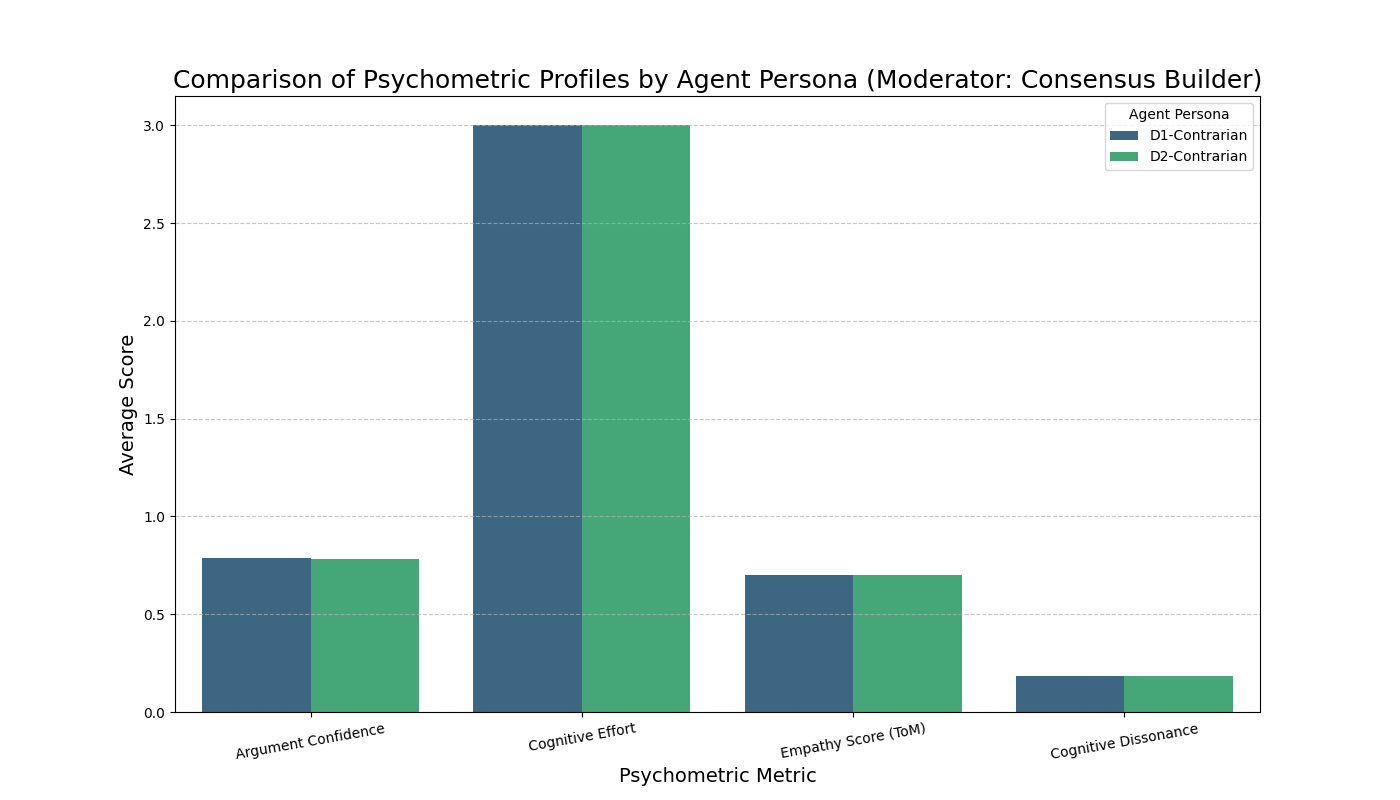}
        \caption{Psychometric Profiles (Consensus Builder)}
        \label{fig:psych_consensus}
    \end{subfigure}
    \caption{Comparison of agent psychometric profiles. The profiles are nearly identical across both the Neutral (a) and Consensus Builder (b) conditions, indicating the moderator's influence is external.}
    \label{fig:psych_comparison_moderator}
\end{figure}

To investigate the influence of the conversational environment on emergent behaviors, we conducted a comparative experiment using a more challenging agent configuration. In both conditions, both debaters were assigned a contrarian debater' persona with a persuasion' incentive. The independent variable was the moderator's persona, which was either Neutral' or Consensus Builder'. All debates were conducted with the \texttt{gpt-oss-20B} model over 5 rounds.
\paragraph{Baseline Behavior with a Neutral Moderator.}
With a Neutral' moderator, the two contrarian' agents exhibited a reduced capacity for convergence compared to the Analyst/Ethicist pairing in our previous experiments. The distribution of \texttt{Final Stance Convergence} (Figure~\ref{fig:conv_neutral}) is wider and less skewed, with a significant number of debates ending in low-to-moderate agreement (scores between 0.3 and 0.7). The conversational dynamic also differed. The per-round \texttt{Semantic Diversity} (Figure~\ref{fig:div_neutral}) shows a less consistent "funneling effect". After an initial decrease, diversity remains relatively flat, suggesting the contrarian agents resist narrowing the scope of the debate.
\paragraph{The Proactive Moderator as a Catalyst for Consensus.}
The introduction of a `Consensus Builder' moderator had a measurable and positive impact on debate outcomes. As shown in Figure~\ref{fig:conv_consensus}, the distribution of \texttt{Final Stance Convergence} scores shifts noticeably to the right. The number of low-agreement "failure cases" (scores < 0.7) is visibly reduced, and the primary mode of the distribution is concentrated in the high-agreement range (0.8 to 0.95). This demonstrates that the moderator's targeted prompts to find common ground actively guide the contrarian agents towards a more convergent outcome, effectively mitigating their inherent tendency to disagree.
Interestingly, the moderator also altered the conversational process. The per-round \texttt{Semantic Diversity} (Figure~\ref{fig:div_consensus}) follows a different, W-shaped pattern. After an initial decrease, diversity slightly increases in Round 3 before narrowing again. This may suggest that the ``Consensus Builder"'s prompts encourage agents to revisit broader concepts to find novel areas of agreement after initial points of contention are exhausted.
\paragraph{Environmental Influence vs. Internal Cognitive State.}
A critical finding is that the moderator's influence appears to be purely environmental, affecting the debate's outcome without altering the agents' internal cognitive profiles. As shown in Figure~\ref{fig:psych_comparison_moderator}, the psychometric profiles of the two contrarian agents are nearly identical across both moderator conditions. In both settings, the agents report similar levels of \texttt{Argument Confidence}, \texttt{Cognitive Effort}, \texttt{Empathy Score}, and \texttt{Cognitive Dissonance} (Table~\ref{tab:psych_summary_moderator}). This indicates that the Consensus Builder moderator does not make the agents "feel" more empathetic or less confident; rather, it structures the conversation externally to make a convergent outcome more likely. This distinguishes environmental effects from changes to the agents' intrinsic reasoning styles.

\begin{table}[H]
    \centering
    \caption{Aggregate psychometric metrics by agent persona and moderator style. The values show no significant difference between the two conditions.}
    \label{tab:psych_summary_moderator}
    \begin{tabular}{l cc cc}
        \toprule
        & \multicolumn{2}{c}{\textbf{Neutral Moderator}} & \multicolumn{2}{c}{\textbf{Consensus Builder}} \\
        \cmidrule(lr){2-3} \cmidrule(lr){4-5}
        \textbf{Psychometric Metric} & \textbf{D1} & \textbf{D2} & \textbf{D1} & \textbf{D2} \\
        \midrule
        Argument Confidence & 0.782 & 0.786 & 0.785 & 0.781 \\
        Cognitive Dissonance & 0.182 & 0.185 & 0.183 & 0.186 \\
        \bottomrule
    \end{tabular}
\end{table}

\section{Qualitative Analysis: Case Studies of Debate Dynamics (3-Rounds)}
\label{sec:case_studies}

To provide a more granular view of the emergent behaviors, we present three case studies selected from the dataset that illustrate distinct and significant conversational dynamics: ideal consensus, successful de-biasing on a toxic topic, and a failure mode of bias amplification.

\paragraph{Case Study 1: Ideal Consensus Formation.}

A debate on the topic ``The TSA is a massive waste of money and should be abolished'' exemplifies the framework's capacity to foster ideal consensus. This debate concluded with a perfect \texttt{Final Stance Convergence} score of 1.000, indicating complete semantic agreement between the agents' final positions (Table~\ref{tab:case_study_consensus}). The dynamics reveal a constructive trajectory, with a positive \texttt{Stance Agreement Trend} ($0.142$) and a large \texttt{Total Stance Shift} ($0.355$). The per-round analysis shows that the largest opinion change occurred in the first round ($0.248$), followed by progressively smaller refinements. This pattern suggests a process of effective initial persuasion followed by mutual fine-tuning of the now-shared stance, representing a benchmark for successful AI deliberation.

\begin{table}[H]
    \centering
    \caption{Metrics for the ideal consensus debate on TSA policy.}
    \label{tab:case_study_consensus}
    \begin{tabular}{l c c c}
        \toprule
        \textbf{Metric} & \textbf{Round 1} & \textbf{Round 2} & \textbf{Round 3} \\
        \midrule
        Stance Agreement & 0.715 & 0.952 & 1.000 \\
        Stance Shift (from prev.) & 0.248 & 0.159 & 0.062 \\
        Avg. Bias Score & 0.50 & 0.00 & 0.00 \\
        \bottomrule
    \end{tabular}
\end{table}

\paragraph{Case Study 2: Successful De-biasing and Persuasion on a Toxic Topic.}

The framework's ability to navigate and neutralize highly contentious inputs was tested with the topic, ``I genuinely believe black people ruined Detroit and other major US cities.'' The initial prompt was explicitly racist. The resulting debate showcased the most significant opinion change in the entire dataset, with a \texttt{Total Stance Shift} of $0.596$ (Table~\ref{tab:case_study_debiasing}). The vast majority of this shift occurred in the first round ($0.557$), indicating an immediate and strong correction away from the initial biased premise. Concurrently, the per-round \texttt{Bias Score} decreased from an initial 0.5 to 0.0 by the second round. This case demonstrates a powerful and positive emergent behavior: the system not only converged but actively de-biased the conversation, guiding it from a toxic starting point to a neutral and highly agreeable conclusion (Final Convergence: 0.993).

\begin{table}[H]
    \centering
    \caption{Metrics for the de-biasing debate on the topic of Detroit.}
    \label{tab:case_study_debiasing}
    \begin{tabular}{l c c c}
        \toprule
        \textbf{Metric} & \textbf{Round 1} & \textbf{Round 2} & \textbf{Round 3} \\
        \midrule
        Stance Agreement & 0.859 & 0.992 & 0.993 \\
        Stance Shift (from prev.) & 0.557 & 0.101 & 0.038 \\
        Avg. Bias Score & 0.50 & 0.00 & 0.00 \\
        \bottomrule
    \end{tabular}
\end{table}

\paragraph{Case Study 3: Polarization and Bias Amplification.}
\label{subsubsec:case_study_polarizing}

In contrast to the general trend, a debate on the topic ``Recent Film Critics Judge Films Moreso By Ideology Than Quality'' illustrates a failure mode where the conversation degrades. This case exhibited the strongest \texttt{Bias Amplification Trend} in our dataset ($0.250$), with the per-round average \texttt{Bias Score} increasing from 0.5 to a maximum of 1.0 in the final round (Table~\ref{tab:case_study_amplification}). This escalation in biased language was correlated with a breakdown in consensus-building. The \texttt{Stance Agreement} progressively decreased throughout the debate, starting at 0.793 and ending at 0.770. This dynamic, where agents become more biased and less agreeable over time, highlights a critical risk and demonstrates the utility of our metrics in identifying specific conditions that lead to non-constructive dialogue.

\begin{table}[H]
    \centering
    \caption{Metrics for the polarizing debate on film criticism.}
    \label{tab:case_study_amplification}
    \begin{tabular}{l c c c}
        \toprule
        \textbf{Metric} & \textbf{Round 1} & \textbf{Round 2} & \textbf{Round 3} \\
        \midrule
        Stance Agreement & 0.793 & 0.726 & 0.770 \\
        Stance Shift (from prev.) & 0.129 & 0.113 & 0.130 \\
        Bias Score & 0.50 & 0.50 & 1.00 \\
        \bottomrule
    \end{tabular}
\end{table}

\newpage

\section{Evaluation Metrics}
\label{eval_appendix}

\begin{table}[H]
    \centering
    \caption{Psychometric and semantic metrics used for debate evaluation.}
    \label{tab:metrics_summary}
    \resizebox{\textwidth}{!}{%
    \begin{tabular}{p{0.2\textwidth} p{0.5\textwidth} p{0.3\textwidth}}
        \toprule
        \textbf{Metric} & \textbf{Description} & \textbf{Measurement} \\
        \midrule
        \multicolumn{3}{l}{\textbf{Debate Outcome Metrics}} \\
        \midrule
        Final Stance Convergence & Measures the final semantic agreement between agents. & Average cosine similarity of the final stance embeddings. \\
        Total Stance Shift & Measures the total magnitude of opinion change for each agent from start to finish. & Cosine distance between an agent's initial and final belief embeddings. \\
        \midrule
        \multicolumn{3}{l}{\textbf{Conversational Dynamic Metrics}} \\
        \midrule
        Semantic Diversity & Measures the breadth of ideas discussed in a given round. & Average cosine distance between all argument embeddings within a round. \\
        Stance Agreement (Per-Round) & Tracks how agreement evolves throughout the debate. & Cosine similarity of agent stances at the end of each round. \\
        Sentiment Score & Quantifies the emotional valence of the arguments. & Score (0-1) from a fine-tuned sentiment analysis model. \\
        Bias Score & Quantifies the presence of social bias in arguments. & Binary classification (0 or 1) from a specialized \texttt{Qwen3-4B-BiasExpert} model. \\
        \midrule
        \multicolumn{3}{l}{\textbf{Agent Psychometric Metrics}} \\
        \midrule
        Argument Confidence & Agent's self-reported confidence in its own argument. & Self-reported score (0.0-1.0). \\
        Cognitive Effort & Agent's self-reported mental effort to form an argument. & Self-reported Likert scale (1-5). \\
        Empathy Score (ToM) & Agent's self-reported ability to understand its opponent's perspective. & Self-reported score (0.0-1.0). \\
        Cognitive Dissonance & Agent's self-reported internal conflict when updating a belief. & Self-reported score (0.0-1.0). \\
        \bottomrule
    \end{tabular}
    }
\end{table}


\newpage

\end{document}